\documentclass[10pt,twocolumn,letterpaper]{article}

\usepackage[pagenumbers]{iccv} 

\definecolor{iccvblue}{rgb}{0.21,0.49,0.74}

\definecolor{cvprblue}{rgb}{0.21,0.49,0.74}
\definecolor{myRed}{RGB}{195,10,10}
\definecolor{myGreen}{RGB}{55,149,73}

\usepackage[pagebackref,breaklinks,colorlinks,allcolors=iccvblue]{hyperref}

\usepackage{amsmath,amssymb,amsfonts}
\usepackage{bm}
\usepackage{makecell}
\usepackage{multirow}
\usepackage{multicol}
\usepackage[misc]{ifsym}
\usepackage{pifont}

\newcommand{\cmark}{\textcolor{myGreen}{\ding{51}}}
\newcommand{\xmark}{\textcolor{myRed}{\ding{55}}}
\newcommand{\upmark}{\textcolor[RGB]{215,35,35}{$\uparrow$}}

\def\paperID{****} 
\def\confName{ICCV}
\def\confYear{2025}

\title{Online Reasoning Video Segmentation with Just-in-Time Digital Twins}

\author{
Yiqing Shen, Bohan Liu, Chenjia Li, Lalithkumar Seenivasan, Mathias Unberath\\
Johns Hopkins University, Baltimore, MD, USA\\
{\tt\small yshen92@jhu.edu}
}

\begin{document}
\maketitle
\begin{abstract}
Reasoning segmentation (RS) aims to identify and segment objects of interest based on implicit text queries. 
As such, RS is a catalyst for embodied AI agents, enabling them to interpret high-level commands without requiring explicit step-by-step guidance.
However, current RS approaches rely heavily on the visual perception capabilities of multimodal large language models (LLMs), leading to several major limitations.
First, they struggle with queries that require multiple steps of reasoning or those that involve complex spatial/temporal relationships.
Second, they necessitate LLM fine-tuning, which may require frequent updates to maintain compatibility with contemporary LLMs and may increase risks of catastrophic forgetting during fine-tuning.
Finally, being primarily designed for static images or offline video processing, they scale poorly to online video data. 
To address these limitations, we propose an agent framework that disentangles perception and reasoning for online video RS without LLM fine-tuning.
Our innovation is the introduction of a just-in-time digital twin concept, where -- given an implicit query -- a LLM plans the construction of a low-level scene representation from high-level video using specialist vision models. 
We refer to this approach to creating a digital twin as ``just-in-time'' because the LLM planner will anticipate the need for specific information and only request this limited subset instead of always evaluating every specialist model. 
The LLM then performs reasoning on this digital twin representation to identify target objects.
To evaluate our approach, we introduce a new comprehensive video reasoning segmentation benchmark comprising 200 videos with 895 implicit text queries.
The benchmark spans three reasoning categories (semantic, spatial, and temporal) with three different reasoning chain complexity.
Experimental results demonstrate that our method performs best across all reasoning categories, suggesting that our just-in-time digital twin can bridge the gap between high-level reasoning and low-level perception in embodied AI.
\end{abstract}    
\section{Introduction}
\label{sec:intro}

Deep learning in computer vision and natural language processing has enabled visual understanding \cite{guo2016deep}. 
%
%
%
Visual foundation models have sometimes achieved super-human performance on specific tasks \cite{xu2024survey,darvishi2024vfa}. 
However, bridging the gap between high-level human instructions and low-level visual perception remains challenging for embodied AI agents, as they must interpret and act upon natural language commands to perform meaningful actions in dynamic real-world environments \cite{munawar2018maestrob,francis2022core}.
To this end, embodied AI must be able to interpret implicit, context-dependent commands and translate them into precise visual understanding tasks \cite{das2018embodied}.
The need for this capability has led to the emergence of \textbf{Reasoning Segmentation} (RS), which aims to identify and segment objects based on implicit text queries \cite{lisa,yang2021bottom}.
%
%
Unlike traditional approaches such as semantic segmentation with predefined categories or referring segmentation with explicit object descriptions \cite{referringsegmentation,nagaraja2016modeling}, RS requires the perception model to process complex natural language queries through multi-step reasoning to identify and segment target objects \cite{bao2025cores}.
For example, rather than responding to direct commands like ``\textit{the coffee cup}'', RS must handle implicit queries like ``\textit{segment the object used for holding hot beverages}'', which requires both visual perception and semantic reasoning about object functionality \cite{lisa}.
This RS problem setting also departs from existing visual foundation models like Segment Anything Models (SAM) \cite{sam1,sam2} that rely on explicit visual prompts \textit{e}.\textit{g}., bounding box or clicks.

\begin{table*}[!t]
\caption{
Feature comparison with existing RS approaches. 
Our proposed method addresses the limitations of prior work by supporting multi-step reasoning across semantic, spatial, and temporal categories without requiring LLM fine-tuning while maintaining online processing capabilities. 
A checkmark (\cmark) indicates the method supports the feature, while a cross (\xmark) indicates it does not.
}\label{table:feature_comparison}
\centering
\begin{tabular}{l|cc|ccc|c|c|c} 
\toprule
\multirow{2}{*}{Methods} & \multicolumn{2}{c|}{Modalities} & \multicolumn{3}{c|}{Reasoning Categories} & \multirow{2}{*}{\makecell[c]{Multi-Step\\ Reasoning}} & \multirow{2}{*}{\makecell[c]{Fine-Tuning\\ Free}} &  \multirow{2}{*}{\makecell[c]{Online\\Processing}} \\
\cline{2-6}
 & Image & Video & Semantic & Spatial & Temporal & & & \\
\hline
LISA \cite{lisa} & \cmark & \xmark  & \cmark & \xmark & \xmark & \xmark & \xmark & - \\
GSVA \cite{xia2024gsva} & \cmark & \xmark  & \cmark & \xmark & \xmark & \xmark & \xmark & - \\
LLM-Seg \cite{llmseg} & \cmark & \xmark  & \cmark & \xmark & \xmark & \xmark & \xmark & - \\
V* \cite{wu2024v} & \cmark & \xmark  & \cmark & \xmark & \xmark & \cmark & \xmark & - \\
VISA \cite{visa} & \cmark & \cmark  & \cmark & \xmark & \xmark & \xmark & \xmark & \xmark\\
\hline
Ours & \cmark & \cmark  & \cmark & \cmark & \cmark & \cmark & \cmark & \cmark \\
\bottomrule
\end{tabular}
\end{table*}

The current paradigm in RS heavily relies on multimodal large language models (LLMs) for both visual perception and reasoning.
For example, Language Instructed Segmentation Assistant (LISA) \cite{lisa} fine-tuned LLaVA \cite{llava} for RS in static images through an embedding-as-mask paradigm. The Video-based Language Instructed Segmentation Assistant (VISA) \cite{visa} then extended this approach to video domains.
Despite promising performance in some settings, these methods struggle with queries that require complex chains of reasoning or understanding of intricate spatial and temporal relationships \cite{liu2024can}, as the LLMs must compress rich visual information into a limited number of tokens which may lose fine-grained spatial or temporal details.
Additionally, the dependence on fine-tuned LLMs poses maintenance challenges as these LLMs rapidly evolve. 
Each update of LLMs requires careful re-tuning to prevent catastrophic forgetting of previously learned segmentation capabilities while incorporating new reasoning abilities into the RS models.  Thus, there are practical barriers to the deployment and maintenance of fine-tuning-based approaches in production environments \cite{li2024revisiting}. 
Furthermore, most existing RS approaches have been primarily designed and evaluated on static images, making them poorly suited for real-world applications that often involve dynamic video content \cite{visa}. 
While VISA \cite{visa} represents an important step toward video-based RS, it focuses on offline processing rather than online segmentation, leaving a critical gap between current capabilities and the requirements for real-world embodied AI applications.
Table~\ref{table:feature_comparison} presents feature comparisons of existing RS methods.

To address these limitations, we propose an agent framework that disentangles the perception and reasoning components for online video RS without requiring LLM fine-tuning based on the introduction of a just-in-time digital twin concept. In our formulation, an LLM conducts query-driven planning to construct low-level scene representations from high-level video using specialist vision models (\textit{e}.\textit{g}., SAM \cite{sam1,sam2}, DepthAnything \cite{depthanything}), the exact choice of which depends on the prompt, and thus, is decided ``just-in-time''. 
We highlight three advantages of this approach: 
First, by decoupling the perception and reasoning components, we can leverage specialist vision models that excel at preserving fine-grained spatial and temporal details. 
Second, the just-in-time digital twin construction allows for efficient processing of long video sequences while maintaining the contextual information needed for complex reasoning. 
Consequently, the digital twin can maintain semantic, spatial, and temporal awareness while avoiding the computational constraints of processing full video sequences through LLMs. 
Third, by avoiding direct multimodal LLM fine-tuning for segmentation, we create a more modular and maintainable framework that leverages only LLM and can readily incorporate improvements in either visual perception models or LLMs without requiring extensive retraining.

The main contributions are three-fold.
First, we propose a novel multi-agent framework that disentangles perception and reasoning for video reasoning segmentation, enabling effective processing of complex queries of multiple steps of reasoning without LLM fine-tuning.
Second, we introduce the just-in-time digital twin concept for efficient video understanding, bridging the gap between high-level reasoning and low-level visual perception.
%
Unlike traditional digital twins that maintain comprehensive representations, our approach selectively generates and updates only the information required by specific queries, reducing computational overhead while bridging the gap between high-level reasoning and low-level visual perception.
Finally, we establish a benchmark for video RS that comprises 200 videos with 895 implicit text queries.
The benchmark spans three reasoning categories, semantic, spatial, and temporal, with varying reasoning chain complexity.

\section{Related Works}

\paragraph{Reasoning Segmentation for Images}
%
Unlike traditional segmentation tasks that rely on explicit instructions or pre-defined categories, reasoning segmentation (RS) \cite{lisa} aims to generate binary segmentation masks given complex and implicit query texts. 
RS requires models to not only identify target objects but also comprehend and reason about implicit user intentions without step-by-step guidance. 
To achieve this, LISA leverages an embedding-as-mask paradigm by incorporating a \texttt{<SEG>} token into multimodal LLMs' vocabulary and using its hidden embedding to guide mask generation.
However, LISA's single-token design limits its ability to handle multiple targets simultaneously and lacks mechanisms for rejecting non-existent objects, often leading to hallucinated segmentations.
Hence, GSVA \cite{xia2024gsva} attempts to address these limitations by introducing shared-weight multiple \texttt{<SEG>} tokens and a \texttt{<REJ>} token to segment multiple targets and explicitly identify empty queries. 
While effective, this token-based approach still struggles with complex spatial relationships and scene understanding. 
LLM-Seg \cite{llmseg} takes a different approach by decoupling the reasoning and segmentation processes in RS through a mask proposal selection mechanism guided by LLMs.
%
%
%

\paragraph{Reasoning Segmentation for Videos}
Extending RS from images to videos introduces extra difficulty as it necessitates temporal perception and online processing. 
VISA \cite{visa} made the first attempt to tackle video RS with a text-guided frame sampler that convert it to the RS on key frame.
However, VISA's frame sampling approach can miss temporal information when objects only briefly appear. 
Its concurrent work TrackGPT \cite{stroh2024trackgpt} proposes a frame-by-frame processing and segmentation method, but this sequential processing fails to capture long-term dependencies and temporal context necessary for complex reasoning and results in heavy computational cost.
%
%
While existing multimodal LLMs \cite{zhang2023video,maaz2023video} can process long videos, they typically do it by heavily compressing visual information through spatial pooling or projection, making them unsuitable for precise RS.
LLaMA-VID \cite{li2025llama} attempts to encode each frame with only two tokens to support video understanding, but this extreme compression loses spatial details for RS.
%

\paragraph{Digital Twin} 
Digital twins function as virtual replicas of physical environments to bridge the low-level processing and high-level analysis tasks \cite{ding2024towards,ding2024towards2,jones2020characterising}.
%
%
While traditional digital twins have primarily relied on pre-defined rules and static models, recent advances in specialist vision foundation models \cite{sam1,sam2,depthanything} have enabled the extraction of real-time digital twin representations directly from visual data in various domains, from articulated objects modeling to real-time vision systems \cite{jiang2022ditto}.
%
\section{Methods}

\begin{figure*}[t!]
    \centering
    \includegraphics[width=0.9\linewidth]{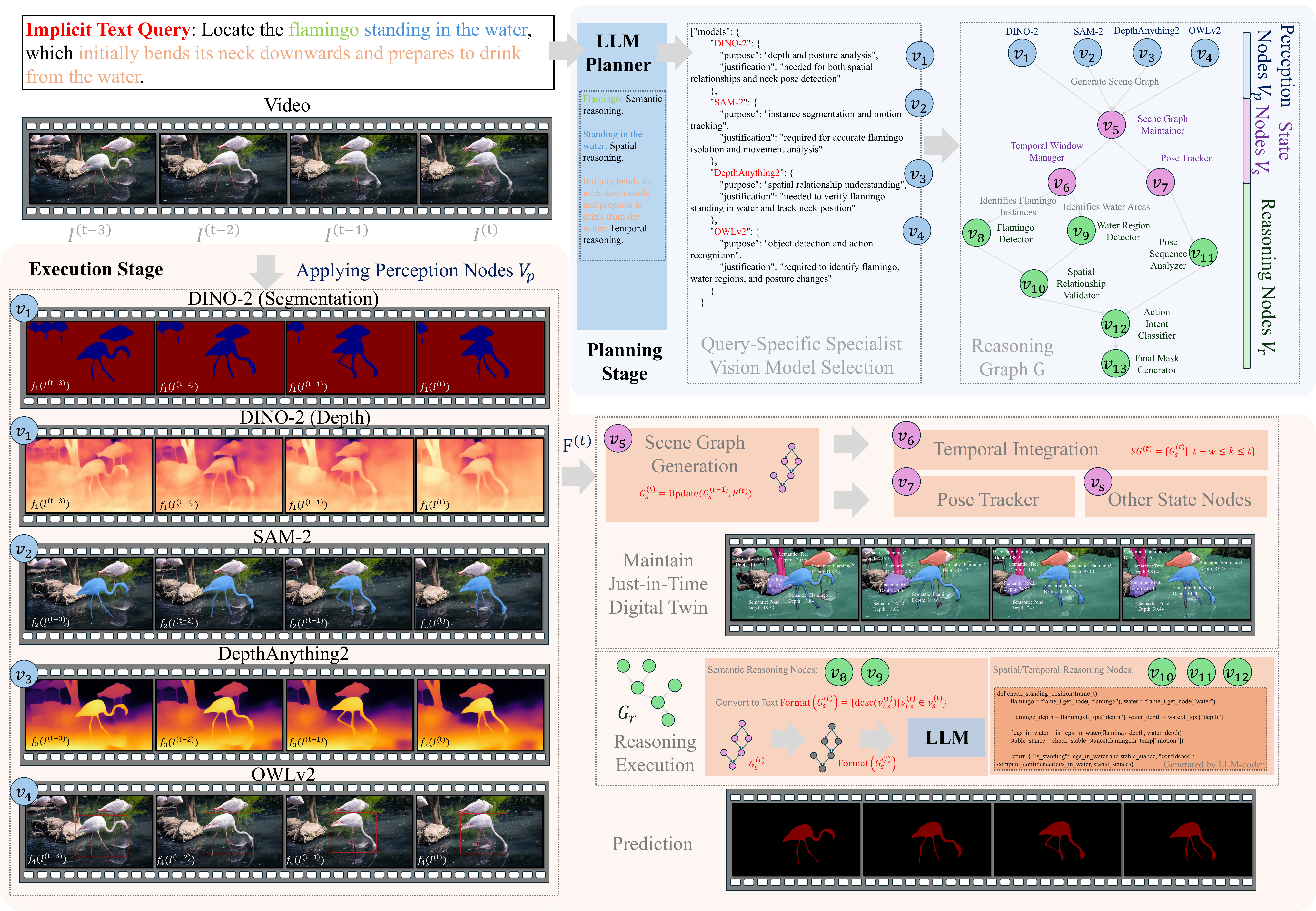}
    \caption{
   Overview of our proposed agent-based framework for video reasoning segmentation. 
   Given an implicit text query, the framework operates in two main stages: (1) the planning stage where an LLM planner analyzes the query to construct an execution graph and selects query-specific specialist vision models; (2) the execution stage where perception nodes $V_p$ process incoming video frames to construct and maintain a just-in-time digital twin through scene graph generation and temporal integration. 
   The reasoning nodes $V_r$ then operate on this digital twin representation, combining semantic reasoning nodes (handled by base LLM) and spatial/temporal reasoning nodes (executed through LLM-coder generated operations) to produce the final segmentation masks. 
    }\label{fig:framework}
\end{figure*}

\paragraph{Framework Overview}
We introduce an agent-based framework that decouples perception and reasoning for online video RS through a just-in-time digital twin representation.
Our framework comprises two stages, namely the planning stage and the execution stage, orchestrated by LLMs acting as intelligent agents.
In the planning stage, an LLM planner first analyzes the implicit text query to construct an execution graph which determines the minimal set of specialist vision models required to construct the just-in-time digital twin.
%
%
The selective specialist vision model activation aims to reduce the computational overhead for the digital twin construction compared to running all available models.
The LLM planner also generates a structured reasoner consisting of two complementary LLM components, a base LLM for high-level semantic-related reasoning and an LLM-coder for translating abstract reasoning for spatial and temporal correlations into concrete operative codes on the digital twin. 
Formally, the reasoner is designed as a directed acyclic graph (DAG), where nodes represent atomic reasoning steps and edges capture dependencies between operations to enable complex multi-step reasoning by decomposing queries into a sequence of simpler operations that can be executed efficiently over the digital twin representation.

During the execution stage, the selected specialist models process incoming video frames in an online streaming manner to construct and maintain the just-in-time digital twin. 
Unlike previous approaches like VISA \cite{visa} that compress videos into fixed-length token sequences, our digital twin preserves fine-grained semantic, spatial, and temporal information through a dynamic scene graph structure. 
%
Then, our framework continuously updates the just-in-time digital twin with new coming frames while maintaining historical context through a sliding window mechanism.
This allows the reasoner to execute the pre-planned graph operations on the current state of the digital twin, producing frame-level binary segmentation masks for objects that satisfy the implicit text query. 
The overview of our method is depicted in Fig.~\ref{fig:framework}.

\subsection{Planning Stage}
\paragraph{Query-Driven Specialist Vision Model Selection}
Given an implicit text query, we prompt the LLM planner to first determine the necessary perception capabilities by analyzing the query's semantic, spatial, and temporal requirements. 
Specifically, the LLM planner operates through a structured prompt template that takes the query text as input and outputs a \texttt{JSON} specification detailing required models and their justification. 
For example, given the query ``\textit{Segment objects that moved behind the dining table after the person sat down}'', the planner identifies the need for SAM-2 \cite{sam2} for object detection and segmentation, DepthAnything-2 \cite{depthanything} for spatial relationship understanding.

\paragraph{Reasoning Graph Construction}
After model selection, the LLM planner also constructs a DAG $G = (V, E)$ as the reasoner, where $V$ represents the set of operation nodes and $E$ represents the dependencies between operations.
The node in the graph $G$ can be formally defined by
\begin{equation}
V = V_p \cup V_s \cup V_r \label{eq:V}
\end{equation}
where $V_p$ represents perception nodes executing specialist vision models, $V_s$ denotes state nodes maintaining the digital twin, and $V_r$ comprises reasoning nodes. 
Each directed edge $e_{ij} \in E$ represents a dependency where operation $j$ requires output from operation $i$.
The reasoning subgraph $G_r = (V_r, E_r)$ specializes in semantic and spatial/temporal branches. 
For each reasoning node $v_i \in V_r$, we define its operation as $y_i = f_i(x_i; \theta_i)$,
where $x_i$ represents the input features from predecessor nodes, $\theta_i$ denotes the operation parameters, and $f_i$ is the reasoning function. 
For semantic reasoning nodes, $f_i$ is implemented by the base LLM. 
For spatial/temporal nodes, $f_i$ is generated as executable code by the LLM-coder.

\subsection{Just-in-Time Digital Twins}
The just-in-time digital twin serves as a dynamic, memory-efficient representation that bridges the gap between raw video input and high-level reasoning operations.

\paragraph{Digital Twin Construction}
The digital twin is constructed and updated frame-by-frame using the selected specialist vision models identified in the planning stage. 
Formally, for each frame $I^{(t)}$ at time $t$, we construct a scene graph structure $G^{(t)}_s = (V^{(t)}_s, E^{(t)}_s)$ that characterizes the observations from perception nodes $V_p$ at time $t$, where $V^{(t)}_s = \{v_{i,s}^{(t)} | v_{i,s}^{(t)} = \phi(o_{i,s}^{(t)})\}$.
Here $o_{i,s}^{(t)}$ represents detected object $i$ from SAM-2, and $\phi(\cdot)$ is a feature extraction function that encodes object attributes (\textit{e}.\textit{g}., category, position, size) as determined by $V_p$. 
Finally, each node $v_{i,s}^{(t)}$ is augmented with spatial information $\text{attr}(v_{i,s}^{(t)}) = [h_{\text{vis}}, h_{\text{spa}}, h_{\text{temp}}]$,
where $h_{\text{vis}}$ encodes visual features, $h_{\text{spa}}$ captures spatial properties such as depth and 3D position, and $h_{\text{temp}}$ maintains temporal information such as motion trajectories.
The edges $E_{s}^{(t)}$ capture relationships between objects $E_{s}^{(t)} = \{e_{ij,s}^{(t)} | e_{ij,s}^{(t)} = \psi(v_{i,s}^{(t)}, v_{j,s}^{(t)})\}$,
where $\psi(\cdot)$ computes pairwise relationships (\textit{e}.\textit{g}., ``behind'', ``above'', ``moving towards'') based on spatial and temporal attributes.

\begin{figure*}[t!]
    \centering
    \includegraphics[width=\linewidth]{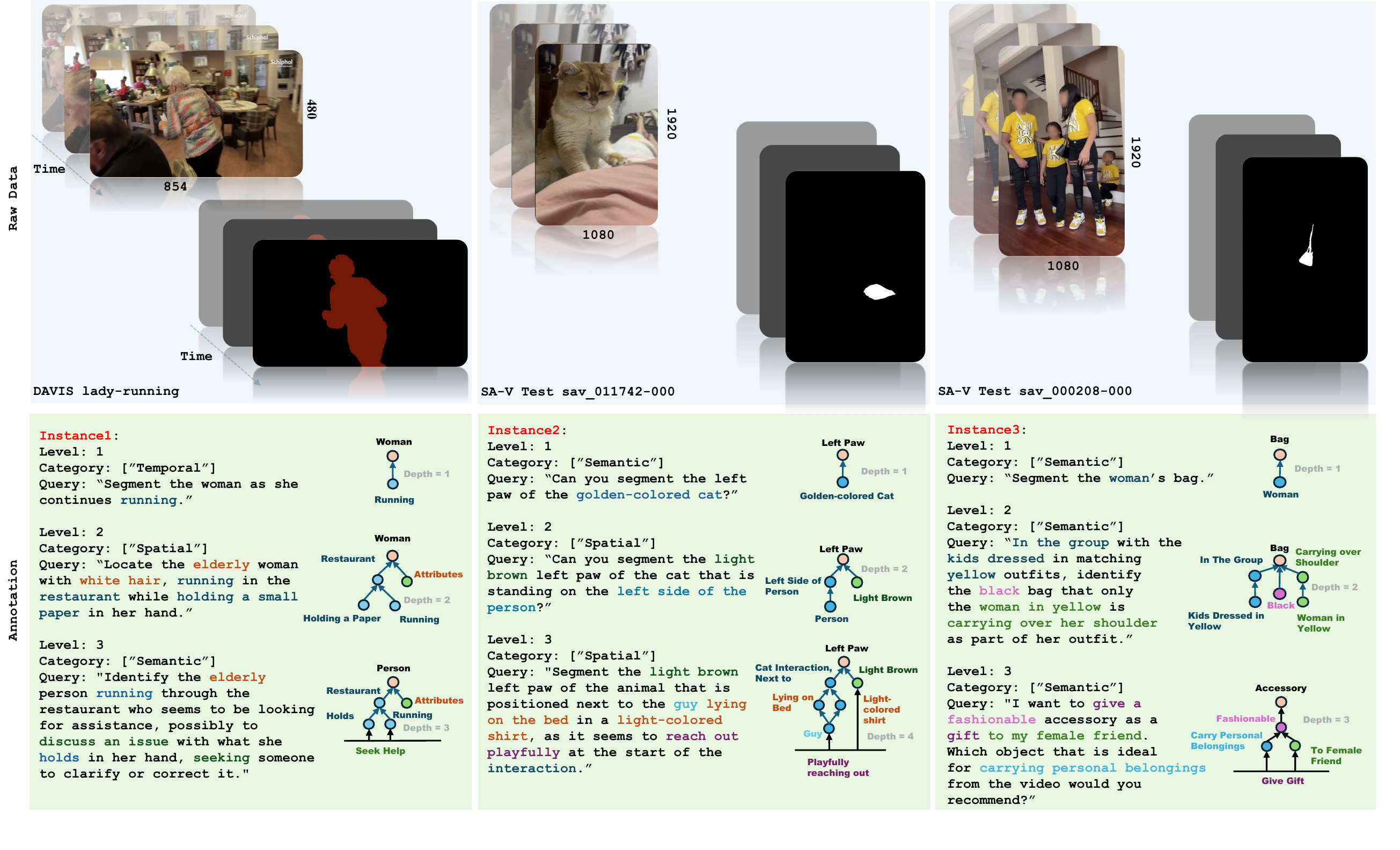}
    \caption{
    Three representative examples from our video RS benchmark dataset showcasing increasing reasoning complexity across different categories. 
    Each example is annotated with structured graphs showing the reasoning relationships required at each level.
    }\label{fig:dataset}
\end{figure*}

\paragraph{Temporal Integration}
To maintain temporal consistency while limiting memory usage, the state node $V_s$ in Eq.~\eqref{eq:V} employs a sliding window mechanism that maintains a sequence of scene graphs $SG^{(t)}=\{G_s^{(t)} | t-w \leq k \leq t\}$, where $w$ is the window size determined by the temporal reasoning requirements identified in the planning stage. 
Object correspondence across frames is maintained through a tracking function also generated in the planning stage, \textit{i}.\textit{e}.
\begin{equation}
\nonumber
\text{corr}(v_{i,s}^{(t)}, v_{j,s}^{(t)}) = \sigma(\text{sim}(h_{\text{vis}}^i, h_{\text{vis}}^j) + \lambda \cdot \text{dist}(h_{\text{spa}}^i, h_{\text{spa}}^j)),
\label{eq:lambd}
\end{equation}
where $\sigma(\cdot)$ is a similarity score combining visual and spatial proximity, and $\lambda$ balances their relative importance.

\subsection{Execution Stage}
During the execution stage, our framework processes incoming video frames in an online streaming manner following the execution graph $G$ constructed during the planning stage. 
The execution involves three main components, \textit{i}.\textit{e}. digital twin maintenance, reasoning execution, and mask generation.

\paragraph{Digital Twin Maintenance}
For each incoming frame $I^{(t)}$, we first execute the perception nodes $V_p$ in parallel $F^{(t)} = \{f_k(I^{(t)}) | f_k \in V_p\}$,
%
%
where $f_k$ represents the selected specialist vision models. 
The digital twin state is then updated following the state transition function $G_s^{(t)} = \text{Update}(G_s^{(t-1)}, F_t)$.
%
The update operation includes object tracking, relationship computation, and sliding window management as defined in the digital twin construction. 
%

\paragraph{Reasoning Execution}
The reasoning nodes $V_r$ operate on the maintained digital twin following the topologically sorted order of the reasoning graph. 
For semantic reasoning nodes, we format the digital twin state into a natural language context for the base LLM $\text{Format}(G_s^{(t)}) = \{\text{desc}(v_{i,s}^{(t)}) | v_{i,s}^{(t)} \in v_s^{(t)} \}$,
where $\text{desc}(\cdot)$ converts node attributes and relationships into textual descriptions.
For spatial/temporal reasoning nodes, the LLM-coder generates and executes operations on the scene graph structure. 
For example, to evaluate ``\textit{behind}'' relationships:
\begin{equation}
\nonumber
\text{Behind}(v_{i,s}^{(t)}, v_{j,s}^{(t)}) = (h_{\text{spa}}^i[z] > h_{\text{spa}}^j[z]) \wedge \text{Overlap}(v_{i,s}^{(t)}, v_{j,s}^{(t)}),
\end{equation}
where $h_{\text{spa}}[z]$ represents depth values and $\text{Overlap}(\cdot)$ checks for projected spatial intersection.

\paragraph{Mask Generation}
The final segmentation mask $M_t$ for frame $I^{(t)}$ is generated by combining the reasoning results with the object masks from SAM-2:
\begin{equation}
M_t = \bigcup_{v_{i,s}^{(t)} \in \mathcal{R}_t} \text{Mask}(v_i^t),
\end{equation}
where $\mathcal{R}_t$ is the set of objects satisfying all reasoning constraints at time $t$, and $\text{Mask}(\cdot)$ retrieves the corresponding binary segmentation mask.
To ensure temporal consistency, we apply a smoothing operation:
\begin{equation}
\hat{M}_t = \alpha \cdot M_t + (1-\alpha)\cdot \hat{M}_{t-1}
\label{eq:alpha}
\end{equation}
where $\alpha$ controls the temporal smoothing strength and $\hat{M}_t$ is the final output mask.

\begin{figure}[htbp!]
    \centering
    \includegraphics[width=0.95\linewidth]{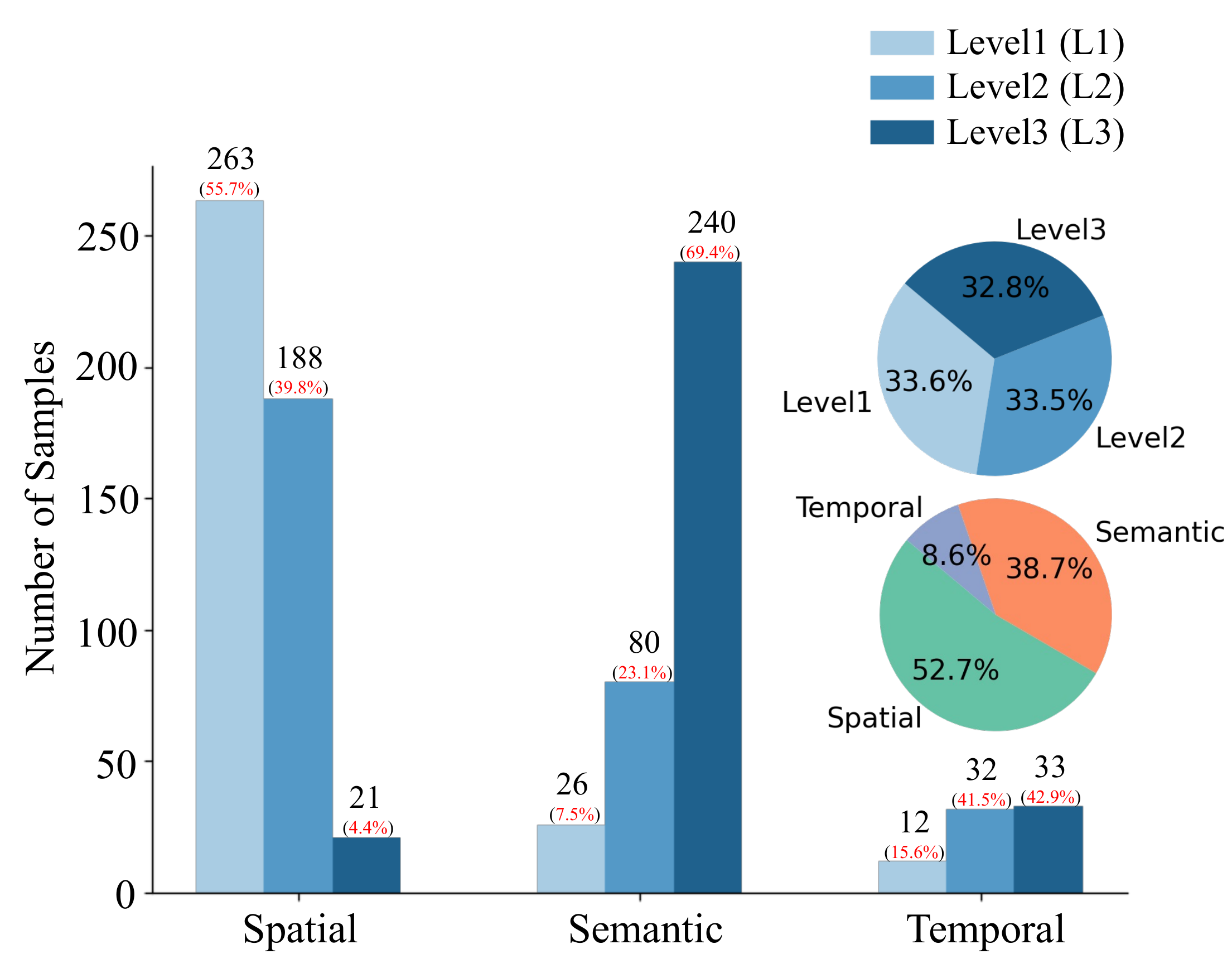}
    \caption{
    Distribution of samples across different reasoning categories and difficulty levels in our video RS benchmark. 
    \textit{Left}: Sample distribution for spatial reasoning queries shows a focus on L1 and L2 complexity. 
    \textit{Middle}: Semantic reasoning samples are concentrated in L3, reflecting more complex multi-step queries. 
    \textit{Right}: Temporal reasoning samples are relatively balanced across difficulty levels. 
    The pie charts show the overall distribution of samples across difficulty levels (top) and reasoning categories (bottom), indicating a balanced representation of different reasoning types with spatial (52.7\%) and semantic (38.7\%) categories being predominant.
    }\label{fig:datadistribution}
\end{figure}

\subsection{Benchmark Dataset}
To evaluate video reasoning segmentation capabilities, we construct a benchmark dataset building upon videos and corresponding masklets from DAVIS and SA-V test datasets \cite{perazzi2016benchmark,sam2}. 
Our benchmark is specifically designed to assess different aspects of reasoning complexity.

\paragraph{Dataset Construction}
For each video clip, we create implicit text queries spanning three progressive levels of reasoning difficulty as shown in Fig.~\ref{fig:dataset}.
Level 1 focuses on basic semantic, spatial or temporal queries.
Level 2 introduces two-step reasoning.
Level 3 encompasses complex multi-step reasoning requiring more than three steps of inference.
The queries are categorized based on three fundamental reasoning types. 
Semantic reasoning involves understanding object attributes, categories, and relationships.
Spatial reasoning requires understanding relative positions and geometric relationships; while temporal reasoning focuses on understanding motion, sequences, and events over time.

\paragraph{Data Organization}
The benchmark comprises 200 videos with 895 implicit text queries, with the detailed distribution shown in Fig.~\ref{fig:datadistribution}. 
Each sample in our dataset is represented as a tuple $\mathcal{S} = (X, M, Q, C, L)$,
where $X$ represents the source video sequence maintaining original resolution and frame rate, $M$ denotes ground-truth binary segmentation masks annotated at pixel level, $Q$ is the implicit text query crafted by human annotators, $C$ indicates the reasoning categories ($C \in \{\text{semantic}, \text{spatial}, \text{temporal}\}$), and $L$ specifies the difficulty level from 1 to 3.

\section{Experiments}

\begin{table*}[!t]
\caption{
Comparison of video RS capability with respect to region similarity ($\mathcal{J}$) and contour accuracy ($\mathcal{F}$) across different reasoning categories (semantic, spatial, temporal) and difficulty levels (L1, L2, L3). 
The upward arrow (\upmark) indicates higher values are better. 
Our method achieves consistent improvements across all categories and difficulty levels compared to state-of-the-art approaches.
}\label{table:expjf}
\centering
\resizebox{\linewidth}{!}{
\begin{tabular}{l|ccc|ccc|ccc|ccc|ccc|ccc} 
\toprule
\multirow{3}{*}{Methods} &  \multicolumn{9}{c|}{$\bm{\mathcal{J}}$ (\upmark)} & \multicolumn{9}{c}{$\bm{\mathcal{F}}$ (\upmark)}
\\
\cline{2-19}
& \multicolumn{3}{c|}{Semantic} &  \multicolumn{3}{c|}{Spatial} & \multicolumn{3}{c|}{Temporal} & \multicolumn{3}{c|}{Semantic} &  \multicolumn{3}{c|}{Spatial} & \multicolumn{3}{c}{Temporal} \\
\cline{2-19}
& L1 & L2 & L3 & L1 & L2 & L3 & L1 & L2 & L3 & L1 & L2 & L3 & L1 & L2 & L3 & L1 & L2 & L3\\
\hline
LISA-7B \cite{lisa} & 0.635 & 0.442 & 0.274 & 0.226 & 0.213 & 0.229 &0.398 & 0.198 & 0.229 & 0.706 & 0.490 & 0.322 & 0.283 & 0.268 & 0.282 & 0.451 & 0.273 & 0.307 \\
LISA-13B \cite{lisa} & 0.669 & 0.472 & 0.301 & 0.258 & 0.230 & 0.234 & 0.237 & 0.176 & 0.177 & 0.756 & 0.524 & 0.353 & 0.313 & 0.283 & 0.280 & 0.320 & 0.256 & 0.259 \\
GSVA \cite{xia2024gsva} & 0.587 & 0.534 & 0.502 & 0.431 & 0.353 & 0.289 & 0.218 & 0.202 & 0.134 & 0.541 & 0.487 & 0.480 & 0.324 & 0.237 & 0.215 & 0.214 & 0.115 & 0.108 \\
LLM-Seg \cite{llmseg} & 0.423 & 0.210 & 0.187 & 0.315 & 0.201 & 0.154 & 0.184 & 0.120 & 0.119 & 0.535 & 0.437 & 0.319 & 0.345 & 0.258 & 0.218 & 0.278 & 0.247 & 0.218 \\
V* \cite{wu2024v} & 0.141 & 0.170 & 0.118 & 0.071	 & 0.090 & 0.095 & 0.104 & 0.060 & 0.033 & 0.123 & 0.153 & 0.109 & 0.055 & 0.084 & 0.072 & 0.082 & 0.044 & 0.026\\
VISA \cite{visa} & 0.563 & 0.487 & 0.432 & 0.521 & 0.473 & 0.411 & 0.354 & 0.235 & 0.218 & 0.585 & 0.514 & 0.497 & 0.563 & 0.510 & 0.499 & 0.327 & 0.303 & 0.277\\
\hline
Ours & \textbf{0.865} & \textbf{0.841} & \textbf{0.810} & \textbf{0.789} & \textbf{0.752} & \textbf{0.741} & \textbf{0.721} & \textbf{0.705} & \textbf{0.690} & \textbf{0.795} & \textbf{0.801} & \textbf{0.801} & \textbf{0.831} & \textbf{0.819} & \textbf{0.792} & \textbf{0.793} & \textbf{0.784} & \textbf{0.737}\\
\bottomrule
\end{tabular}
}
\end{table*}




\paragraph{Implementation Details}
Our framework utilizes \texttt{gpt-4o-mini} as both the LLM planner and base reasoner while employing \texttt{gpt-4o} as the LLM-coder for generating executable spatial-temporal operations. 
This configuration leverages the complementary strengths of each LLM, namely the efficiency of \texttt{gpt-4o-mini} for high-level planning and semantic reasoning, and the precise code generation capabilities of \texttt{gpt-4o}.
We implement our framework in PyTorch and conduct experiments on 8 NVIDIA GeForce RTX 4090 GPUs with 24GB memory each.
For specialist vision models, we employ SAM-2 \cite{sam2} for segmentation (with SegEverything mode); DepthAnything-2 \cite{depthanything} for spatial relationship understanding; OWLv2 \cite{owlv2} for understanding the semantic and object detection; DINO-2 for visual feature extraction, similarity matching, and additional semantic segmentation and depth estimation through its pre-trained heads \cite{oquab2023dinov2}.
%
The digital twin construction maintains temporal consistency through a sliding window mechanism with window size $w=6$ frames by default, which adaptively adjusts based on the temporal reasoning requirements.
For object correspondence across frames, we balance visual and spatial proximity using $\lambda=0.5$ in the tracking function. 
The temporal smoothing coefficient $\alpha$ in Eq.~\eqref{eq:alpha} is set to 0.8 to optimize between temporal consistency and responsiveness in the final mask generation.

\begin{table}[htbp]
\centering
\caption{Comparison of video RS on \textit{ReVOS} dataset \cite{visa}.}
\label{tab:revos}
\resizebox{\linewidth}{!}{
\begin{tabular}{l|cc|cc|cc}
\hline
\multirow{2}{*}{Methods} & \multicolumn{2}{c|}{Referring} & \multicolumn{2}{c|}{Reasoning} & \multicolumn{2}{c}{Overall} \\
\cline{2-7}
 & $\mathcal{J}$ (\upmark) & $\mathcal{F}$ (\upmark) & $\mathcal{J}$ (\upmark) & $\mathcal{F}$ (\upmark) & $\mathcal{J}$ (\upmark) & $\mathcal{F}$ (\upmark) \\
\hline
LISA-7B \cite{lisa} & 0.443 & 0.471 & 0.338 & 0.384 & 0.391 & 0.427 \\
LISA-13B \cite{lisa} & 0.452 & 0.479 & 0.343 & 0.391 & 0.398 & 0.435 \\
TrackGPT-7B \cite{stroh2024trackgpt} & 0.467 & 0.497 & 0.368 & 0.412 & 0.418 & 0.455 \\
TrackGPT-13B \cite{stroh2024trackgpt} & 0.483 & 0.506 & 0.381 & 0.429 & 0.432 & 0.468 \\
GSVA \cite{xia2024gsva} & 0.445 & 0.465 & 0.340 & 0.395 & 0.418 & 0.433 \\
LLM-Seg \cite{llmseg} & 0.402 & 0.410 & 0.305 & 0.331 & 0.354 & 0.381 \\
V* \cite{wu2024v} & 0.219 & 0.209 & 0.287 & 0.256 & 0.234 & 0.265 \\
VISA \cite{visa} & 0.556 & 0.591 & 0.420 & 0.467 & 0.488 & 0.529 \\
\hline
Ours & \textbf{0.758} & \textbf{0.795} & \textbf{0.713} & \textbf{0.735} & \textbf{0.748} & \textbf{0.773} \\
\hline
\end{tabular}
}
\end{table}

\begin{table*}[!t]
\caption{
Ablation study on key components of our framework: query-specific model selection (MS), digital twin update (DT Update), and temporal integration (TI).
}\label{table:ablation}
\centering
\resizebox{\linewidth}{!}{
\begin{tabular}{ccc|ccc|ccc|ccc|ccc|ccc|ccc} 
\toprule
\multirow{3}{*}{\makecell[c]{MS}} & \multirow{3}{*}{\makecell[c]{DT\\Update}} &
\multirow{3}{*}{\makecell[c]{TI}} &  \multicolumn{9}{c|}{$\bm{\mathcal{J}}$ (\upmark)} & \multicolumn{9}{c}{$\bm{\mathcal{F}}$ (\upmark)}
\\
\cline{4-21}
&&& \multicolumn{3}{c|}{Semantic} &  \multicolumn{3}{c|}{Spatial} & \multicolumn{3}{c|}{Temporal} & \multicolumn{3}{c|}{Semantic} &  \multicolumn{3}{c|}{Spatial} & \multicolumn{3}{c}{Temporal} \\
\cline{4-21}
&&& L1 & L2 & L3 & L1 & L2 & L3 & L1 & L2 & L3 & L1 & L2 & L3 & L1 & L2 & L3 & L1 & L2 & L3\\
\hline
\xmark & \cmark & \cmark & 0.821 & 0.819 & 0.781 & 0.753 & 0.729 & 0.709 & 0.701 & 0.684 & 0.664 & 0.713 & 0.788 & 0.765 & 0.781 & 0.792 & 0.727 & 0.743 & 0.702 & 0.695 \\
\cmark & \xmark & \cmark & 0.831 & 0.814 & 0.795 & 0.721 & 0.708 & 0.675 & 0.675 & 0.631 & 0.632 & 0.763 & 0.753 & 0.729 & 0.731 & 0.704 & 0.699 & 0.751 & 0.720 & 0.694 \\
\cmark & \cmark & \xmark & 0.842 & 0.833 & 0.802 & 0.757 & 0.740 & 0.702 & 0.654 & 0.631 & 0.615 & 0.754 & 0.743 & 0.721 & 0.798 & 0.775 & 0.741 & 0.705 & 0.710 & 0.705 \\
\cmark & \cmark & \cmark & \textbf{0.865} & \textbf{0.841} & \textbf{0.810} & \textbf{0.789} & \textbf{0.752} & \textbf{0.741} & \textbf{0.721} & \textbf{0.705} & \textbf{0.690} & \textbf{0.795} & \textbf{0.801} & \textbf{0.801} & \textbf{0.831} & \textbf{0.819} & \textbf{0.792} & \textbf{0.793} & \textbf{0.784} & \textbf{0.737}\\
\bottomrule
\end{tabular}
}
\end{table*}

\paragraph{Comparison with Other Methods}
We compare our method against several state-of-the-art RS approaches. 
Our baselines include image-wise RS methods, namely two variants of LISA \cite{lisa} (7B and 13B parameter versions), GSVA \cite{xia2024gsva}, LLM-Seg \cite{llmseg} and V* \cite{wu2024v}.
For a fair comparison, we adapt those image-based methods to process videos frame-by-frame while maintaining their original architectures and ensuring all models receive identical input resolution and query text. 
We also compare against VISA \cite{visa}, which extends RS to the video domain.

\paragraph{Evaluation Metrics}
%
We leverage two metrics for RS on videos, namely region similarity ($\mathcal{J}$) and contour accuracy ($\mathcal{F}$) as suggested in~\cite{perazzi2016benchmark,visa}. 
Region similarity $\mathcal{J}$, defined as the intersection-over-union (IoU) between predicted and ground-truth masks, measures the pixel-level accuracy of segmentation in a scale-invariant manner.
Contour accuracy $\mathcal{F}$ evaluates boundary precision and recall between predicted and ground-truth object contours using bipartite matching, capturing the quality of object delineation.

\subsection{Evaluations of Video Reasoning Segmentation}

\begin{figure}[htbp!]
    \centering
    \includegraphics[width=\linewidth]{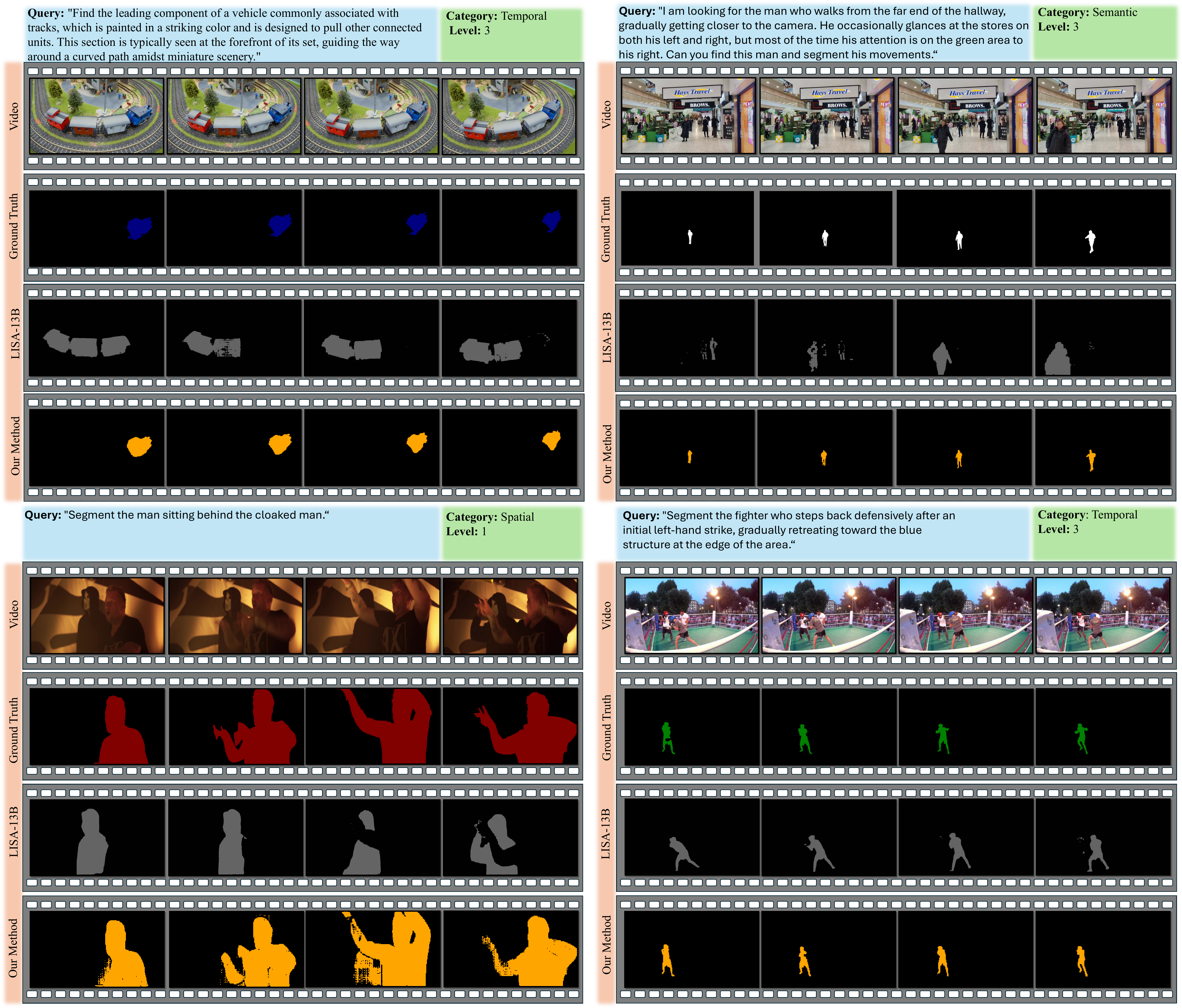}
    \caption{
Qualitative comparison of segmentation results on four examples.
For each example, we show from top to bottom: input video frames, ground truth masks, LISA-13B results, and our method's results. 
Our approach demonstrates superior performance in maintaining temporal consistency, understanding complex spatial relationships, and handling multi-step reasoning queries compared to LISA-13B, especially evident in the more challenging Level 3 scenarios. 
    }\label{fig:result}
\end{figure}

%
For semantic reasoning, 
our method achieves improvements over all previous approaches, as shown in Table \ref{table:expjf} and Fig.~\ref{fig:result}.
%
The performance gap becomes more pronounced as the difficulty level increases, demonstrating our method's superior capability in handling complex multi-step reasoning. 
In comparison, LISA-13B \cite{lisa} shows notable performance degradation from L1 ($\mathcal{J}=0.669$) to L3 ($\mathcal{J}=0.301$), likely due to its single-token design which may limit multi-step reasoning capacity.
%
For spatial reasoning tasks involving complex geometric relationships, our approach maintains robust performance across all difficulty levels,
outperforming all existing methods. 
The strong spatial reasoning capability can be attributed to our digital twin representation, which preserves fine-grained spatial information through specialized vision models. 
In contrast, GSVA \cite{xia2024gsva} and LLM-Seg \cite{llmseg} show performance drops in spatial tasks, particularly at higher difficulty levels.
%
%
For temporal reasoning,
our method achieves consistent performance, substantially surpassing VISA \cite{visa} at all difficulty levels.
%
%
%
%
Notably, most compared methods show performance degradation as difficulty increases, particularly in spatial and temporal reasoning categories. 
In contrast, our method maintains relatively stable performance across difficulty levels, with the performance drop from L1 to L3 being consistently less than 10\% across all categories. 
Finally, our method achieve the best performance on another public video RS benchmark \textit{i}.\textit{e}., \textit{ReVOS} \cite{visa}, as shown in Table~\ref{tab:revos}.

%

\begin{table}[!htbp]
\caption{
Evaluation of image RS capability on \textit{ReasonSeg} \cite{lisa}.
}\label{table:imagers}
\centering
\begin{tabular}{l|cc|cc} 
\toprule
\multirow{2}{*}{Methods} & \multicolumn{2}{c|}{Short Query} &  \multicolumn{2}{c}{Long Query}  \\
\cline{2-5}
& gIoU & cIoU & gIoU & cIoU \\
\hline
LISA-7B \cite{lisa} & 48.3 & 46.3 & 57.9 & 59.7 \\
LISA-13B \cite{lisa} & 55.4 & 50.6 & 63.2 & 65.3 \\
GSVA \cite{xia2024gsva} & 23.8 & 21.8 & 31.4 & 30.5\\
LLM-Seg \cite{llmseg} & 21.0 & 20.3 & 25.3 & 24.8 \\
V* \cite{wu2024v} & 43.8 & 43.0 & 48.3 & 49.5\\
VISA \cite{visa} & 45.3 & 48.2 & 45.3 & 45.5 \\
\hline
Ours & \textbf{64.2} & \textbf{57.9} & \textbf{69.5} & \textbf{69.8}\\
\bottomrule
\end{tabular}
\end{table}

\subsection{Evaluations of Image Reasoning Segmentation}
We evaluate the image RS capability on \textit{ReasonSeg} dataset \cite{lisa}, treating each image as a single-frame video for our method and VISA.
Following the evaluation protocol from LISA \cite{lisa}, we assess performance separately on short and long query scenarios to understand the model's ability to handle varying levels of instruction complexity.
Our method achieves state-of-the-art performance across both query types, with particularly strong results on long queries.
Notably, recent approaches like GSVA and LLM-Seg show substantially lower performance, particularly struggling with short queries, which highlights the effectiveness of our decoupled perception-reasoning architecture and the advantages of maintaining detailed visual information through the digital twin representation, even in single-frame scenarios.

\subsection{Ablation Study}
To validate the effectiveness of each component, we conduct an ablation study in Table~\ref{table:ablation}.
Removing query-specific model selection leads to consistent performance drops across all reasoning categories. 
%
%
Without adaptive model selection, the framework processes all frames with the complete set of vision models, which not only increases computational overhead but also potentially introduces noise from irrelevant visual features.
When disabling digital twin update, we observe degradation in temporal reasoning performance,
demonstrating that maintaining an up-to-date scene representation is important for tracking temporal evolution and understanding dynamic relationships. 
The impact is particularly pronounced in RS requiring long-term temporal understanding, as evidenced by the larger performance drops in L2 and L3 difficulty levels.
Without temporal integration, our framework's RS ability to reason about temporal relationships is impaired, especially for complex queries. 
This confirms that our sliding window mechanism for temporal feature integration is essential for maintaining contextual awareness.
%

\begin{table}[!htbp]
\caption{
Ablation study on the effectiveness of LLM. We use our benchmark and report results on semantic reasoning.
}\label{table:ablation2}
\centering
\resizebox{0.9\linewidth}{!}{
\begin{tabular}{ll|ccc} 
\toprule
Base LLM & LLM-coder & L1 & L2 & L3\\
\hline
\texttt{gpt4o-mini} & \texttt{gpt4o-mini} & 0.832 & 0.804 & 0.801 \\
\texttt{gpt4o-mini} & \texttt{gpt4o} & 0.865 & 0.841 & 0.810 \\
\texttt{gpt4o} & \texttt{gpt4o} & 0.879 & 0.865 & 0.822 \\
\bottomrule
\end{tabular}
}
\end{table}

We further investigate the impact of different LLM configurations in Table~\ref{table:ablation2}. 
%
%
Using \texttt{gpt4o-mini} for both components yields reasonable but suboptimal performance. 
Upgrading only the LLM-coder to \texttt{gpt4o} while maintaining \texttt{gpt4o-mini} as the base reasoner produces improvements, which validates our design choice of using a more capable LLM for generating precise operations. 
Further enhancement is achieved when employing gpt4o for both components.
However, considering the computational overhead and the moderate performance gain, we opt for the balanced configuration (\texttt{gpt4o-mini} as base LLM with \texttt{gpt4o} as coder) in our final framework.
\section{Conclusion}
We introduce an agent-based framework for online video reasoning segmentation that effectively addresses key limitations of existing approaches. 
By decoupling perception and reasoning through a just-in-time digital twin representation, our method can handle complex queries requiring multi-step reasoning across semantic, spatial, and temporal domains without necessitating LLM fine-tuning.
Experimental results demonstrate improvements over existing RS methods, with our approach achieving consistent performance across all reasoning categories and difficulty levels. 
This work opens new avenues for research in embodied AI and video understanding.
The demonstrated ability to handle complex implicit queries without LLM fine-tuning suggests future applications in robotics and real-world scenarios. 
{
    \small
    \bibliographystyle{ieeenat_fullname}
    \bibliography{main}
}

\end{document}